\documentclass[sigconf]{acmart}

\usepackage[utf8]{inputenc}

\renewcommand{\footnotetextcopyrightpermission}[1]{}
\usepackage{booktabs}
\usepackage{subcaption}

\newcommand{\lcs}{SupRB}
\newcommand{\lcsone}{SupRB-1}

\newcommand{\XX}{\mathcal{X}}
\renewcommand{\AA}{\mathcal{A}}

\newcommand{\Xeval}{X_{\text{eval}}}
\newcommand{\xdim}{{D_\XX}}
\newcommand{\adim}{{D_\AA}}
\newcommand{\XA}{\{X, A\}^T}
\newcommand{\xa}{\{x, a\}^T}
\newcommand{\XAtrain}{\{X, A\}^T_\text{train}}
\newcommand{\XAvalid}{\{X, A\}^T_\text{valid}}
\newcommand{\qtrain}{q_\text{train}}

\newcommand{\RR}{\mathbb{R}}
\newcommand{\amax}{a_{\text{max}}}
\newcommand{\amaxp}{\hat{a}_{\text{max}}}
\newcommand{\amaxpc}{\hat{a}_{\text{max}_c}}
\newcommand{\qp}{\hat{q}}
\newcommand{\TT}{\mathtt{T}}
\newcommand{\FF}{\mathtt{F}}

\newcommand{\xat}[1][a]{\{x, {#1}\}^T}
\newcommand{\q}[1][a]{q(x, a)}

\begin{document}
\title{\lcs: A Supervised Rule-based Learning System for Continuous Problems}

\author{Michael Heider}
\authornote{These authors contributed equally to the paper.}
\orcid{0000-0003-3140-1993}
\affiliation{%
  \institution{University of Augsburg}
  \department{Organic Computing Group}
  \streetaddress{Eichleitnerstr.\@ 30}
  \city{Augsburg}
  \country{Germany}
  \postcode{86159}
}
\email{michael.heider@informatik.uni-augsburg.de}

\author{David P\"atzel}
\authornotemark[1]
\orcid{0000-0002-8238-8461}
\affiliation{%
  \institution{University of Augsburg}
  \department{Organic Computing Group}
  \streetaddress{Eichleitnerstr.\@ 30}
  \city{Augsburg}
  \country{Germany}
  \postcode{86159}
}
\email{david.paetzel@informatik.uni-augsburg.de}

\author{J\"org H\"ahner}
\affiliation{%
  \institution{University of Augsburg}
  \department{Organic Computing Group}
  \streetaddress{Eichleitnerstr.\@ 30}
  \city{Augsburg}
  \country{Germany}
  \postcode{86159}
}
\email{joerg.haehner@informatik.uni-augsburg.de}

\renewcommand{\shortauthors}{Michael Heider et al.}

\begin{abstract}
We propose the \lcs{} learning system, a new accuracy-based Pittsburgh-style learning classifier system (LCS) for supervised learning on multi-dimensional continuous decision problems.
\lcs{} learns an approximation of a quality function from examples (consisting of situations, choices and associated qualities) and is then able to make an optimal choice as well as predict the quality of a choice in a given situation.
One area of application for \lcs{} is parametrization of industrial machinery.
In this field, acceptance of the recommendations of machine learning systems is highly reliant on operators' trust.
While an essential and much-researched ingredient for that trust is \emph{prediction quality}, it seems that this alone is not enough.
At least as important is a \emph{human-understandable explanation} of the reasoning behind a recommendation.
While many state-of-the-art methods such as artificial neural networks fall short of this, LCSs such as \lcs{} provide human-readable rules that can be understood very easily.
The prevalent LCSs are not directly applicable to this problem as they lack support for continuous choices.
This paper lays the foundations for \lcs{} and shows its general applicability on a simplified model of an additive manufacturing problem.
\end{abstract}

\keywords{Learning Classifier Systems, Evolutionary Machine Learning, Manufacturing}

\maketitle

\section{Introduction}
\label{sec:Introduction}

Parametrization of industrial machinery is often determined by human operators.
These specialists usually obtained most of their expertise through year-long experimental exploration based on prior knowledge about the system or process at play.
Transferring that knowledge to other operators with as little loss as possible (e.\,g.\@ to new colleagues whenever experienced operators retire or to end users of the machinery after commissioning is finished) is a challenge:
Humans' ability of exactly attributing parameter choices to the situations that led to them and then communicating this knowledge in a comprehensible manner tends to be rather restricted---which leads to new operators being forced to repeat exploration to learn for themselves.
Machine learning (ML) can help with this, for example, by supporting new operators or users with recommendations or simply by recording existing experiences and extracting knowledge to make it available at a later point.

Parts of an operator's knowledge can be seen as a collection of mappings from parametrizations for the machine and variables beyond their influence to an expected process quality resulting from them---abstractly speaking, a collection of if-then rules with outcomes subject to noise.
While many ML methods represent knowledge in a less or differently structured way, this is not the case for \emph{learning classifier systems} (LCSs) whose models are collections of human-readable if-then rules constructed using ML techniques and model structure optimizers \cite{Holland1976, Urbanowicz2009}.
This learning scheme is thus suited naturally to incorporate an operator's knowledge as externally specified rules can be included directly.
Also, due to their inner structure, LCSs can more easily provide explainations for their predictions.
Due to this transparency towards human users, compared to black box systems, an increased trust by operators that contained knowledge and thus recommendations are correct can be expected; which is essential for these system's actual applicability.

This paper proposes the \lcs{} learning system, a new accuracy-based Pittsburgh-style LCS for supervised learning on continuous multi\--dimension\-al decision problems such as the one of parametrization of industrial machinery.
Pittsburgh-style LCSs \cite{Smith1980} have a model structure optimizer (in classic Pittburgh-style systems, a genetic algorithm (GA)) operate on a population of rule collections of variable length each of which represents a potential solution to the learning problem at hand.

This work focuses on solving the problem of parametrization optimization of industrial machinery, which is defined in Section \ref{sec:ParameterOptimization}.
An LCS architecture that solves this problem, \lcs{}, is introduced in Section \ref{sec:GeneralLCS} along with its first implementation \lcsone{} in Section \ref{sec:Implementation}.
\lcsone{} is evaluated on different function approximation problems in Section \ref{sec:Evaluation}.
Section \ref{sec:relatedwork} gives an account of related research.

\section{Parametrization optimization}
\label{sec:ParameterOptimization}

Parametrization optimization is the process of finding the best parameter choice, or \emph{parametrization}, for a given system $S$ with regard to some \emph{quality measure} $q$.
One such parametrization can be viewed as a vector $a \in \AA \subseteq \RR^\adim$ where $\AA$ is the parametrization space, $\adim$ is the number of parameters to be optimized and each component of $a$ corresponds to one adjustable system parameter for $S$.
Which parametrization is optimal regarding $q$ depends on a number of environmental factors (e.\,g.\@ ambient temperature or humidity) in addition to characteristics of process, machine, material and the part to be produced.
For a given system $S$, we call one instance of those additional factors a \emph{situation}; situations can again be assumed to be represented by a vector $x \in \XX \subseteq \RR^\xdim$ where $\XX$ is called the situation space and $\xdim$ is the fixed dimensionality of situations for $S$.
Having defined parametrizations and situations, we can now specify the quality measure's form as
\begin{equation}
  \label{eq:QualityMeasure}
  q : \{\XX, \AA\}^T \to \RR
\end{equation}
where every $\xat = (x_1, \dots, x_\xdim, a_1, \dots, a_\adim)^T \in \{\XX, \AA\}^T$ is a stacked vector consisting of a situation $(x_1, \dots, x_\xdim) \in \XX$ and a parametrization $(a_1, \dots, a_\adim)\in \AA$.
For readability, we write $q(x, a)$ instead of $q(\xa)$.
The target of $q$ is a single scalar which is possibly derived appropriately from a vector of multiple quality features.
We assume that $q(x, a)$ is at least continuous in $a$ which we think is realistic in most real-world scenarios:
\begin{equation}
  \label{eq:QContinuous}
  \lim_{a \to a_0} q(x, a) = q(x, a_0)
\end{equation}
With the definition for $q$, we can now define the optimization problem that describes the search for an optimal parametrizations for a given situation $x$:
\begin{align}
  \label{eq:OptimizationProblem}
  \underset{a}{\text{maximize }} & q(x, a)
\end{align}
Note that, realistically, neither $q$ nor its derivative can be assumed to be known (albeit either of those would simplify the problem greatly).
Instead, we assume that the only information about $q$ is a fixed set of examples.

Thus, the learning problem we consider is:
Given a fixed set of $N$ examples for situations and parametrizations $\{\{x, a\}^T\}$ as well as their respective qualities $\{\q{}\}$, learn to predict for a given unknown situation $x \in \XX$ a parametrization $\amaxp{}(x) \in \AA$ for which
\begin{equation}
  \label{eq:Prediction}
  \amaxp{}(x) \approx \amax{}(x) = \underset{a}{\text{argmax }} q(x, a)
\end{equation}
where $\amax{}(x)$ is the actual optimal parametrization in situation $x$.

A natural way of measuring improvements on this learning problem is the following:
A model can be said to be an improvement over another on a set of situations $\Xeval{} \subset X$ if the \emph{actual quality} of the \emph{predicted} optimal parametrizations on those situations is closer to the \emph{actual quality} of the \emph{actual} optimal parametrizations.
This can be quantified, for example, by using the mean error for an error measure $L$ on the model's prediction:
\begin{equation}
  \label{eq:ModelPerformance}
  \frac{1}{|\Xeval|} \sum_{x \in \Xeval} L(q(x, a_{\text{max}}(x)), q(x, \hat{a}_{\text{max}}(x)))
\end{equation}

\section{An LCS architecture for continuous problems}
\label{sec:GeneralLCS}

This section presents a high-level view of \lcs{}, the overall LCS architecture we propose, in order to solve parametrization optimization problems which were introduced in the previous section.

\subsection{Model structure}
\label{sec:ModelStructure}

Just like other LCSs, \lcs{} forms a \emph{global} model from a population $C$ of \emph{local} models, called \emph{classifiers}; in the case of \lcs{}, the global model is meant to approximate the quality measure $q$ defined in Section \ref{sec:ParameterOptimization}.
Each classifier is responsible for a subspace of the input space $\XX$; which subspace it is for a certain classifier is specified by that classifier's \emph{condition} which is also sometimes called its \emph{localization}.
The set of classifier conditions forms the overall model structure of \lcs{}; this structure fulfills a similar role as the graph structure of a neural network in that it needs to be chosen carefully in order for the system to perform well.
At that, performing well is not just about approximating $q$ as close as possible; there are usually additional goals such as being explainable (cf.\@ Section \ref{sec:Introduction}) which require the model structure's complexity to be as low as possible.

\subsection{Local models: Classifiers}
\label{sec:GeneralClassifiers}

A classifier $c$ consists of three main components:

\begin{itemize}
\item
  Some representation of a matching function $m_c : \XX \to \{\TT{}, \FF{}\}$.
  We say that $c$ \emph{matches} situation $x$ iff $m_c(x) = \TT$.
  Correspondingly, we say that $c$ \emph{does not match} $x$ iff $m_c(x) = \FF$.
\item
  Some local model approximating $q$ on all $x \in \XX$ which the classifiers matches.
\item
  An estimation of the classifier's goodness-of-fit on the situations it matches (solely used in classifier mixing).
\end{itemize}

Be aware that the classifiers' matching functions' domain is $\XX$ and not $\{\XX, \AA\}^T$.
This increases explainability greatly as an ideal partitioning of $\XX$ (total, without overlaps) entails that there is exactly one rule regarding the parametrization for each possible situation.
Conversely, if we partitioned in $\{\XX, \AA\}^T$ optimally, there would possibly still be multiple rules for a given situation as the system might have partitioned in the dimensions of $\AA$ as well.
Since we assume continuity of $q(x, a)$ regarding $a$ (see (\ref{eq:QContinuous})), partitioning in $\AA$ would only be necessary if the local models could not capture $q$'s behaviour in $\AA$, for example because it is highly multi-modal.
In that case, partitioning in $\AA$ might be sensible, as would using more sophisticated local models.

\subsection{Epoch-wise training}
\label{sec:Epochs}

Training an LCS can generally be divided into two subproblems:
For once, the classifiers' local models need to be trained so that the predictions they make on the subspace they are responsible for are as accurate as possible.
Secondly, the overall model structure of the LCS has to be optimized: the classifier's localizations have to be aligned in such a way that every local model can capture the characteristics of the subspace it is assigned to as well as possible.

Michigan-style LCS such as XCS(F) \cite{butz2002, wilson2002a} or ExSTraCS \cite{urbanowicz2015a} try to solve these problems incrementally by, for each seen example, performing a single update on some of the classifier's local models and then improving these classifier's localizations.
This approach is especially sensible when learning has to be incremental (e.\,g.\@ in reinforcement learning settings).
However, due to the learning problem being non-incremental (all training data is available from the very start), \lcs{} can be trained non-incrementally.
This means that training can be done in two separate phases that are repeated alternatingly until overall convergence \cite{drugowitsch2007b}, each phase being responsible for solving one of the subproblems:
\begin{enumerate}
\item (Re-)train each local classifier model on the data that it (now) matches.
\item Optimize the model structure (i.\,e.\@ the set of classifier conditions), for example using a heuristic such as a GA.
\end{enumerate}
At that, each phase is executed until it converges or some termination criterion, such as a fixed number of updates, is met.
It is important to note that during fitting of each phase's parameters, the parameters of the respective other are considered fixed---otherwise, convergence cannot be guaranteed.
If a GA is used for the model structure optimization, then this GA works on a population of classifier populations---these kind of systems are commonly called Pittsburgh-style LCS.
However, since we expect many optimization methods to be applicable to this (see Section \ref{sec:FutureWork}), an implementation of \lcs{} does not necessarily contain a GA.
Nevertheless, the general \lcs{} architecture should probably be placed into or close to the Pittsburgh-style category.
The \emph{implementation} of \lcs{} we present in Section \ref{sec:Implementation} is definitely a Pittsburgh-style LCS since it uses a GA to optimize the model structure.

Dividing the learning process into two distinct phases is advantageous.
First of all, optimization of the process's hyperparameters can be done more straightforwardly because hyperparameters are divided into two disjoint sets, one for each of the two phases.
Besides that, the learning process is analysed more easily because the overall optimization problem of fitting the model to the data decomposes nicely into the two subproblems solved by the phases \cite{drugowitsch2007b}---`nicely' meaning, that solving the subproblems independently of each other solves the overall problem.
For example, if learning does not work and, upon inspection, the classifier weight updates converge correctly and fast enough, it is immediatly clear that the model structure optimization is the culprit and corresponding measures can be taken.

\subsection{Prediction}

After training, \lcs{} can make two kinds of prediction.
A \emph{quality prediction} consists of using \lcs{}'s internal function approximation to predict the quality resulting from a certain parametrization $a$ given a certain situation $x$.
To do so, \lcs{} retrieves all classifiers from the classifier population that match $x$, that is, the set
\begin{equation}
  M(x) = \{c \in C \mid m_c(x) = \TT\}.
\end{equation}
The predictions of these classifiers then need to be \emph{mixed} in order to yield the overall \emph{system prediction} for the inputs, $\qp(x, a)$.
One way of mixing is a simple sum which is weighted by some accuracy measure $F_c$ defined for each classifier $c$:
\begin{equation}
  \label{eq:MixingGeneral}
  \qp(x, a) = \sum_{c \in M(x)} F_c \qp_c(x, a)
\end{equation}
Here, $\qp_c(x, a)$ denotes the quality value that the local model of $c$ predicts for parametrization $a$ in situation $x$.
It is important that
\begin{equation}
  \label{eq:NormalizedWeights}
  \sum_{M(x)} F_c = 1
\end{equation}
or otherwise the classifier's combined predictions systematically over- or undershoot the actual value as the local models must be trained independently \cite{drugowitsch2007b}, which means that they are unaware of the other local model's predictions during training.

A \emph{parametrization choice} (or $\amaxp$-prediction) consists of predicting the best parametrization for a given fixed situation $x_0$, that is, performing (\ref{eq:OptimizationProblem}) to yield (\ref{eq:Prediction})---which is the more central kind of prediction for parametrization optimization.
The way of doing this highly depends on the used form of local models.
For example, if the local models are polynomial functions of a degree of less than five, an exact analytical solution exists (Abel-Ruffini theorem) as partial derivatives can be used to find the set of local optima $A_{\text{local}}$ from which \lcs{} then can retrieve the global optimum by using its function approximation:
\begin{equation}
  \amaxp = \underset{a \in A_{\text{local}}}{\text{argmax }} \qp(x_0, a)
\end{equation}
For other functions, where an exact analytical solution is unknown or impractical, there are other options that range from root-finding algorithms \cite{brent1973} to heuristics such as hill climbing with random restarts \cite{russel2009}, genetic algorithms \cite{holland1975} or chemical reaction optimization \cite{lam2010}.
Although these non-analytical methods require a comparably larger amount of computation time, they are feasible in the setting \lcs{} targets:
Industrial processes that are being optimized are usually preplanned anyway, which takes a lot longer than any of the heuristics needs to find \lcs{}'s parametrization choice.

\section{\lcsone{}: A first implementation of \lcs{}}
\label{sec:Implementation}

While the previous section introduced \lcs{}'s general architecture, learning process as well as its desired prediction capabilities, we now want to give a detailed account of \lcsone{}, a first implementation of that system\footnote{Which we will make available in the camera-ready version.}.

\subsection{Training and validation sets}
\label{sec:Data}

\lcsone{} randomly splits the available training data, $\XA$, into two disjoint sets of configurable sizes, $\XAtrain \sqcup \XAvalid$, a training and a validation set.
The training set is used exclusively to fit the classifier's local models to the data they match whereas the validation set is used exclusively to optimize the model structure.
This approach is rather simplistic; incorporating more sophisticated sample management (k-fold cross validation etc.) is planned for the future.

To simplify representation and computation, we assume that parametrization and situation values are normalized to $[-1, 1]^\adim$ and $[-1, 1]^\xdim$, respectively; this means that $\AA \simeq \AA_\text{actual}$ needs to hold where $\AA_\text{actual}$ is the \emph{actual} action value that is reported back to an external system.
Given the context of optimizing parameters of industrial machinery it is reasonable to assume that upper and lower bounds for $\AA_{actual}$ exist in all cases which makes this normalization trivial.

\subsection{Classifiers}

Classifier conditions are interval-based using an ordered bound representation \cite{wilson2001, stone2003}; an extension to hyper-ellipsoids \cite{butz2005c} is already in the works.

All classifiers' local models are a simple \emph{linear regression}\footnote{We use the one from the Python library \emph{scikit-learn} \cite{scikit-learn}.} on a subset of the \emph{second order polynomial features} of the input which is fitted on $\XAtrain$ and $\qtrain$.
In order to be able to analytically derive the $\amaxp$-prediction, we exclude all combinations of different dimensions of $\AA$ resulting in the following features set:
\begin{equation}
  \label{eq:Features}
  \{ x_ix_j, x_ia_k, a_k^2 \mid i, j \in 1, \dots, \xdim, k \in 1, \dots, \adim \}
\end{equation}
The reasoning behind our choice for second order polynomial features instead of linear models alone is that a linear model's maximum is always at one of the boundaries of the domain, if they exist.

The classifiers' goodness-of-fit is measured using a \emph{mean squared error} on $\XAtrain$.
We don't use a separate validation set for estimating the goodness-of-fit in order to be as sample efficient as possible; in the industrial machinery context this system mainly targets, labeled data sets are comparably small.
Nevertheless, a separate validation set for goodness-of-fit estimation would most certainly help the system to evolve better generalizing solutions faster.

\subsection{GA for optimizing the model structure}

We optimize the model structure (the classifier's localizations) using a simple GA whose population consists of classifier populations.
As already stated above, this makes \lcsone{} a Pittsburgh-style LCS.

The GA is \emph{generational} with a configurable number of \emph{elitists} (cf.\@ for example \cite{holland1975}); a steady-state version was implemented as well but in a few short preliminary tests did not perform significantly better (albeit there is still no conclusive answer yet).
The GA performs \emph{mutation} and \emph{crossover} on classifier populations.

For a single classifier population, mutation changes the bounds of the interval-based conditions of all classifiers by a normal distribution widened by a step-size $s$.
Mutation steps are clipped at the minimum lower and maximum upper interval bounds, $-1$ and $1$ respectively (this can be disabled via a hyperparameter) in order to keep the hyperrectangle described by the classifier's condition entirely within $[-1, 1]^\xdim$ (confer Section \ref{sec:Data}).
Given any bound (lower or upper) $b \in \XX$, its mutated value is distributed according to
\begin{equation}
  \label{eq:Mutation}
  \max(1, \min(-1, \{b + s * \mathcal{N}(\mu = 0, \sigma = 1)\}))
\end{equation}
The step size $s$ is initially set to one thousandth of the maximum interval width, which is $\frac{2}{1000}$ in our normalized case.
\lcsone{} adapts $s$ by the well-known one-fifth rule as it is used in \cite{auger2009} with a small update factor of $F = 1.05$.

Crossover is done similarly as in \cite{drugowitsch2007b} but using a normal distribution instead of a uniform one in order to keep offspring sizes closer together and less often generating really small offspring.
 Given two parents of lengths $l_1$ and $l_2$, a number $l_1'$ is drawn from $\mathcal{N}(\mu = l_1 + l_2 / 2, \sigma = 1)$ repeatedly until $1 \leq l_1' \leq l_1 + l_2 - 1$ (the condition ensures that each offspring contains at least one classifier).
 After that, the classifiers of both parents are shuffled together and divided randomly among two children, one of size $l_1'$, the other of size $l_1 + l_2 - l_1'$.
Performing this kind of crossover too often might be too disruptive making a crossover rate necessary.

\lcs{} selects parents for crossover using a simple \emph{tournament selection} with tournaments of size $2$ and the individual with the highest fitness always winning \cite{miller1995}.
We measure fitness relatively between two individuals $i_1$ and $i_2$ based on their respective mean squared errors $e_1$ and $e_2$ on the validation set $\XAvalid$ and their respective lengths $l_1$ and $l_2$ which is a na\"ive measure for their model structure's complexity.
Individual $i_1$ wins the tournament if either of the following is true:
\begin{equation}
  \label{eq:Fitness1}
  e_1 < e_2 \quad \wedge \quad l_1 \leq \frac{e_2}{e_1} l_2
\end{equation}
or
\begin{equation}
  \label{eq:Fitness2}
  e_1 \geq e_2 \quad \wedge \quad l_1 \leq k \frac{e_2}{e_1} l_2
\end{equation}
where $k \in [0, 1]$ is a hyperparameter weighing higher solution complexities against lower errors.
This means that if $i_1$'s error on $\XAvalid$ is smaller than $i_2$'s, $i_1$'s complexity is allowed to be up to $\frac{e_2}{e_1}$ ($> 1$) times larger than the one of $i_2$.
On the other hand, in order for $i_1$ to win the tournament with a higher error, it needs to have an at least $k \frac{e_2}{e_1}$ (< 1) times smaller complexity than $i_2$.

\subsection{Initialization}
\label{sec:Initialization}

The GA's population of classifier populations is initialized by randomly generating a number of individuals of a user-specified fixed size.
We experimented with initializing randomly-sized individuals up to an upper bound to have a higher chance of finding the `correct' individuals' size early on.
However, that did not (yet) work out---it seems that an initially larger overall amount of classifiers is more important than finding the correct solution size quickly.

Classifiers for an individual are generated randomly by sampling the bounds of their match function's intervals uniformly from $\XX$.
Although this is one of the least sophisticated methods and results in a high chance of initial overlaps and unmatched examples, it seemed to result in far larger overall stability when compared to initializing classifier populations with evenly spaced individuals.
The reason is probably the greater genetic diversity in the system.

At the end of initialization, all classifiers of all classifier populations are fitted to the examples from $\XAtrain$ that they match once.

\subsection{Fitting classifiers}

\lcs{} uses the most simple linear regression model from \emph{scikit-learn} \footnote{\texttt{sklearn.linear\_model.LinearRegression}} as the local model for each classifier.
These models provide a builtin means of fitting them to data which we use with standard parametrization, which minimizes the $L^2$-norm by Ordinary Least Squares.

\subsection{Prediction}

Due to the simplicity of the classifier's local models, \lcsone{} can easily perform the two kinds of prediction the \lcs{} architecture postulates.
To predict a quality value given a situation and an action, $\xa$, the linear regression models of all classifiers $c$ that match $x$ are queried for their respective predictions $\qp_c(\xa)$.
These predictions are then mixed with weights based on the classifiers' normalized goodness-of-fit $g$ which we calculate from their mean squared error on $\XAtrain$.
The unnormalized goodness-of-fit of a single classifier is:
\begin{equation}
  \label{eq:GoodnessOfFit}
  g'_c =
  \frac{1}{
    \frac{e_c + 1}{
    E}
  }
\end{equation}
where $E = \sum_{c' \in M(x)} e_{c'} + 1$ is the sum of the errors of all classifiers matching $x$ and serves as normalization term for the error.
Note the---for now---na\"ive addition of $1$ to all errors in order to avoid zero terms.
We further have to normalize $g'$ again, in order to fulfill (\ref{eq:NormalizedWeights}) yielding
\begin{equation}
  g_c = \frac{g'_c}{G'}
\end{equation}
where $G' = \sum_{c' \in M(x)} g'_{c'}$ is the sum of the unnormalized goodness-of-fit values of the classifiers matching $x$.
Substituting into (\ref{eq:MixingGeneral}) results in the following mixing model:
\begin{equation}
  \label{eq:QMixing}
  \qp(x, a) = \sum_{c \in M(x)} g_c \qp_c(x, a)
\end{equation}

An parametrization choice $\amaxp(x)$ for a given situation $x$ again results from mixing each matching classifier $c$'s prediction.
At that, $\amaxpc(x)$ can be derived analytically based on the implicit paraboloids that performing a linear regression on a second order polynomial feature space yields.
For example,
\begin{itemize}
\item if the paraboloid opens downwards, the parametrization choice is the position of the vertex in $\AA$ whereas
\item if the paraboloid opens upwards, it is one of the points in
  \begin{equation}
    \label{eq:APrediction}
    \{ (a_1, \dots, a_\adim) \mid a_1, \dots, a_\adim \in \{-1, 1\} \}.
  \end{equation}
\end{itemize}
The parametrization choices of the matching classifiers are then mixed using the same procedure based on their respective mean squared errors on $\XAtrain$ which gives
\begin{equation}
  \label{eq:AMixing}
  \amaxp(x) = \sum_{c \in M(x)} g_c \amaxpc(x).
\end{equation}

\subsection{Summary of hyperparameters}
\label{sec:Hyperparameters}

We now want to give a quick overview of all the hyperparameters introduced so far and a discussion of their impact.

Test set size is the percentage of the training data used exclusively for evaluating the individuals' fitness (see Section \ref{sec:Data}).
This value is only really critical whenever there is only little data available as in that case, a trade-off has to be made between giving more data to the process of fitting local model predictions versus the problem structure optimization.
By incorporating more sophisticated training data organisation techniques, however, this hyperparameter loses some if not most of its impact.

The size of the initial individuals corresponds to the initial number of classifiers in the system (see Section \ref{sec:Initialization}).
Having enough genetic diversity from the start is extremely relevant, so a higher value is generally better.
However, higher values naturally tend to increase the time until a compact solution is found.

The GA's population size is a less sensitive hyperparameter as long as there exist enough classifiers at the very start.
Higher values allow the system to explore more search space in the same number of generations while generations themselves need more computation time.
A similar argument goes for the number of elitists: While it is certainly important to have some amount of elitism for most problems in order to not accidentally forget good solutions, the actual amount of elitists in the population seems not to be that relevant.

Last, but not least, $k$ (see (\ref{eq:Fitness1}) and (\ref{eq:Fitness2})) is the most impactful hyperparameter as it directly interferes with the used fitness measure and thus with the evolutionary pressures within the system.
A higher value (closer to 1) emphasizes the generation of less complex solutions while allowing for a higher error.
In the long-term, this value should probably made dependant on the dimensionalities of the input space, $\xdim$ and $\adim$, because in higher-dimensional spaces, a slight deviation from the intended target can lead to comparably larger errors than in spaces of lower dimensionality.
We expect this effect to lead to different behaviour for the same $k$ on problems that only differ in their dimensionality but not in their general characteristics.
However, we defer a closer look at this to future work.

The other hyperparameters have not that high of an impact and are discussed more in-depth in the publications referenced at their first mention. Table \ref{fig:Hyperparameters} gives a quick reference of all hyperparameters, their expected impact as well as a proposed default value.

\begin{table*}[t]
  \centering
  \begin{tabular}{l l l}
  hyperparameter             & impact                        & default\\
  \hline
  validation set size              & usually low                   & 0.5\\
  individuals' initial sizes & medium to high                & 30\\
  GA population size         & usually low                   & 30\\
  number of elitists         & depends on GA population size & 1\\
  $k$ (fitness parameter)    & high                          & too problem dependent\\
  $F$ (one-fifth rule parameter) & low                       & 1.05 \\
  crossover rate & medium & 0.9 (more results pending) \\
\end{tabular}
\caption{Overview of hyperparameters of \lcsone{} and their default value.}
\label{fig:Hyperparameters}
\end{table*}

\section{Evaluation}
\label{sec:Evaluation}

We evaluated \lcsone{} on two computable problems which are discussed together with the obtained results in the following sections.

\subsection{Frog Problem}

The 2-dimensional frog problem \cite{wilson2004} was already used in the evaluation of systems with similar goals as \lcs{}, namely GCS \cite{wilson2007}, XCSFCA \cite{tran2007} and XCSRCFA \cite{iqbal2012} (cf.\@ Section \ref{sec:relatedwork} for more information on these systems), which is why it was chosen for this work as well.
Essentially a reinforcement learning problem with episode length $1$ and continuous states, actions and rewards, achieving maximal performance constitutes choosing an action which is equal to the situation.
\begin{equation}
  P(x,a)= 
  \begin{cases}
    x+a,     & \text{if } x+a \leq 1\\
    2-(x+a), &  \text{if } x+a \geq 1
  \end{cases}
\end{equation}

We trained and evaluated \lcsone{} with standard parameter settings (cf.\@ Table \ref{fig:Hyperparameters}) and $k = 0.1$.
After 100 generations with only 50 training and 50 validation examples, the fitness elitist was able to consistently (averaged over 30 runs) reach an MSE of less than $0.05$ on choosing the optimal action when given states from a holdout evaluation set.
Regarding predicting the quality of a state-action pair the MSE was below $0.03$ on the same data.
The fitness elitist of the final generation contained 36 classifiers.

GCS, XCSFCA and XCSRCFA all evaluate the function 100,000 times, which is considerably more than \lcsone{}'s 100 evaluations which took place to generate the training data. 
However, this direct comparison is slightly unfair as the three other systems could probably also have used a sample far smaller than 100,000 with a similar training procedure:
They showed system errors below 0.05 after only 10,000 evaluated samples.
Nevertheless, given the few examples required for training \lcsone{}, a high sample efficiency is very likely.
GCS stagnates at an error of 0.05 with about 1400 classifiers, while XCSFCA achieves an error below 0.01 after 30,000 samples with 740 classifiers whereas XCSRCFA solves the problem perfectly after 18,000 samples using about 740 classifiers \cite{iqbal2012}.
It should be noted that \lcsone{} achieves a much greater rule compactness, while definitely performing worse in terms of overall function approximation error.

It can be assumed that the higher function approximation error originates in the fact that \lcsone{} does not partition the search space in $\AA$ (see Section \ref{sec:GeneralClassifiers}) and therefore has to fit the non-continuous function with paraboloids, which can not achieve a perfect approximation performance.

\subsection{AM-Gauss}

The AM-Gauss problem is a simplified model of an FDM-based additive manufacturing (AM) process's part quality and was created using expert knowledge.
The process itself consists of material (usually thermoplastic polymers) being melted and then extruded to gradually construct a part whose quality depends on a number of factors such as the temperature to which the material is heated.
For a given material (one dimension of $\XX$), the resulting part quality varies at increasing temperatures:
Up until the melting point any resulting part's quality can be expected to be zero as no part can be produced at all. 
With a further increase in temperature, quality tends to increase as well at a rate depending on material properties up until some---unknown---point where the material becomes too soft to remain in shape which degrades part quality.
At even higher temperatures, material might simply fail to successfully construct the part at all at which point quality can effectively be treated as zero again.
This relationship of material, temperature and resulting quality can be simplified to a Gaussian function.

The FDM-based AM process we consider contains five continuous (obviously a simplification by itself) situation dimensions:
Material, printer, room temperature, humidity and the kind of part to produce.
These situations interact with six continuous parameters:
Extrusion temperature, print bed temperature, cooling fan speed, extruder movement speed, material retraction speed and retraction distance (the first four parameters are rather self explanatory; the latter two come into effect whenever the extruder can not construct the part using continuous movement and has to move without extruding material).
Assuming that every combination of situation dimensions and parameters can be modeled by a Gaussian function as motivated above leads to the following overall model for the quality function:
\begin{equation}
  \label{eq:AM-Gauss}
  q(y) =
  q\left(\begin{pmatrix}x_1\\\vdots\\x_5\\a_1\\\vdots\\a_6\end{pmatrix}\right) =
  \sum_{\substack{j \in 1, \dots, 11,\\k \in 1, \dots, 11,\\k \neq j}}
  \exp\left({-\left(\begin{pmatrix}y_j\\y_k\end{pmatrix}-s\right)^T P_{j,k} \left(\begin{pmatrix}y_j\\y_k\end{pmatrix}-s\right)}\right)
\end{equation}
Here, the $P_{j,k}$'s are randomly generated positive semi-definite matrix in $\RR^{2x2}$ with eigenvalues in $[0, 30]$ (ensures sensible scaling) and $s$ is a randomly generated vector in $[-1, 1]^2$ representing the shift of the Gaussian function.
We did not include noise in our model, however, an evaluation on more realistic noisy environments is already planned.

\begin{figure*}[h]
  \begin{subfigure}[t]{.3\textwidth}
    \includegraphics[width=\linewidth]{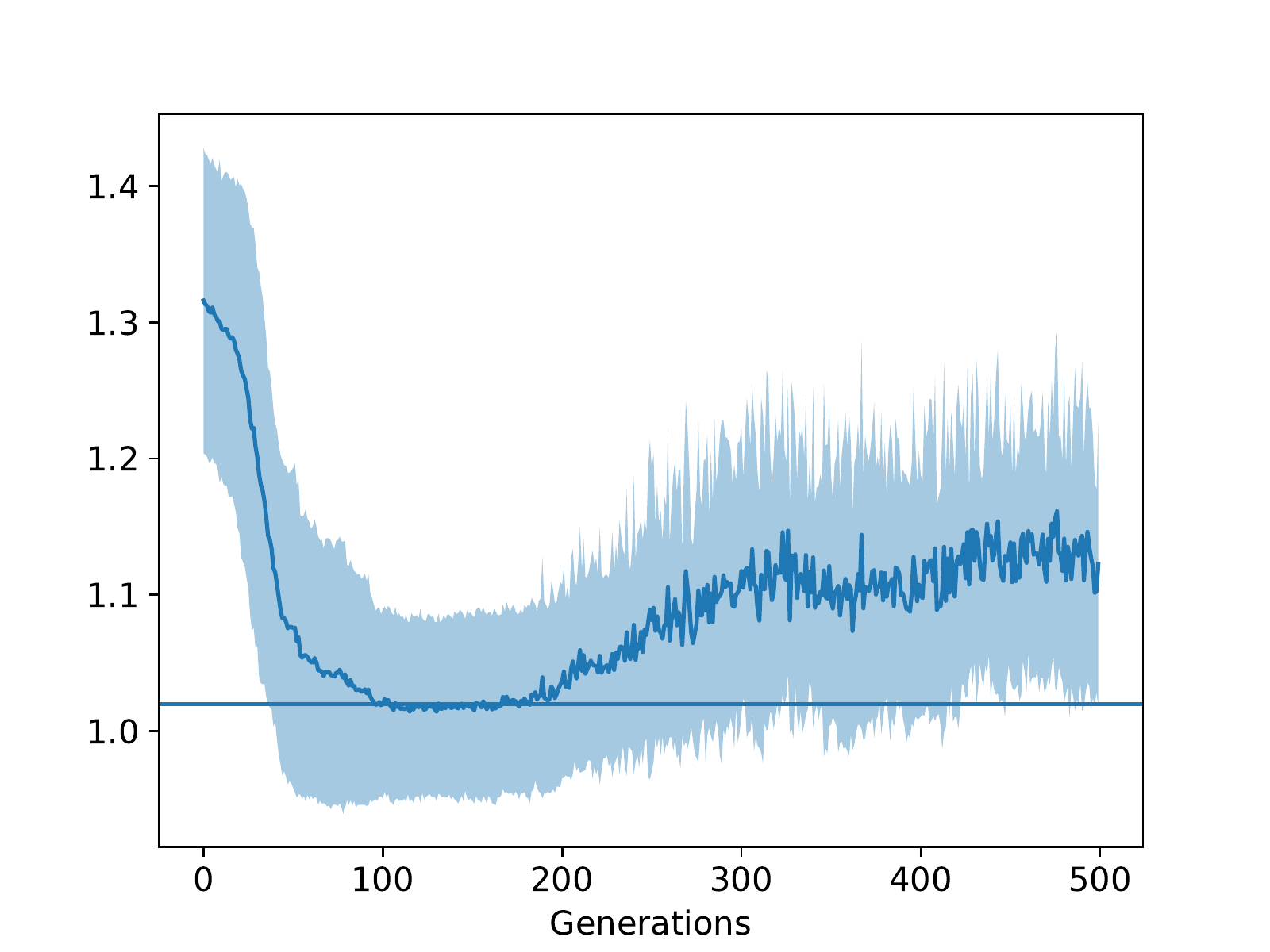}
    \caption{
      \emph{Quality predictions} on holdout data with ANN baseline.
    }
    \label{fig:RMSEFitEval}
  \end{subfigure}
  \hfill
  \begin{subfigure}[t]{.3\textwidth}
    \includegraphics[width=\linewidth]{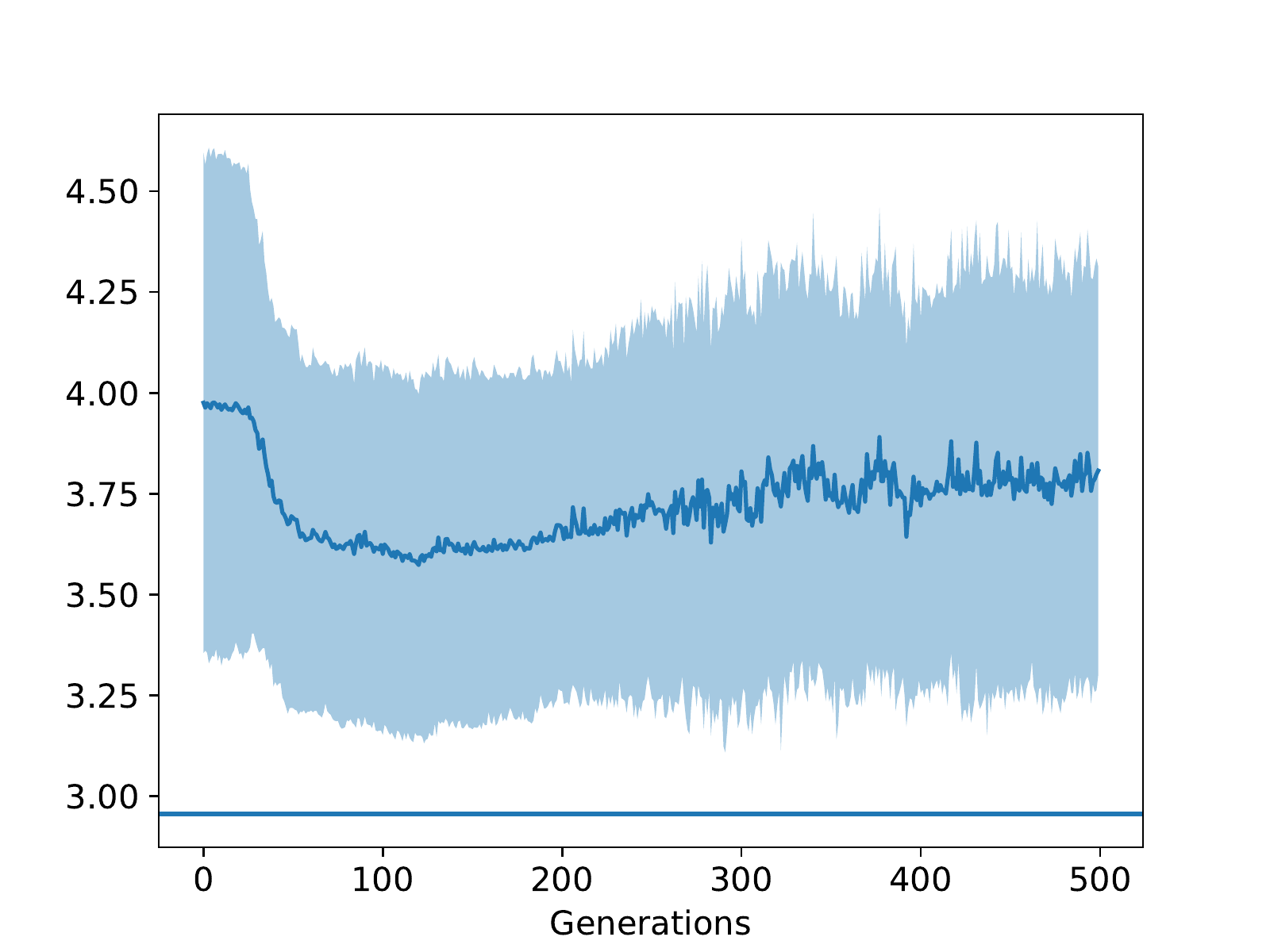}
    \caption{
      \emph{Parametrization choices} on holdout data with ANN baseline.
    }
    \label{fig:RMSEOptimum}
  \end{subfigure}
  \hfill
  \begin{subfigure}[t]{.3\textwidth}
    \includegraphics[width=\linewidth]{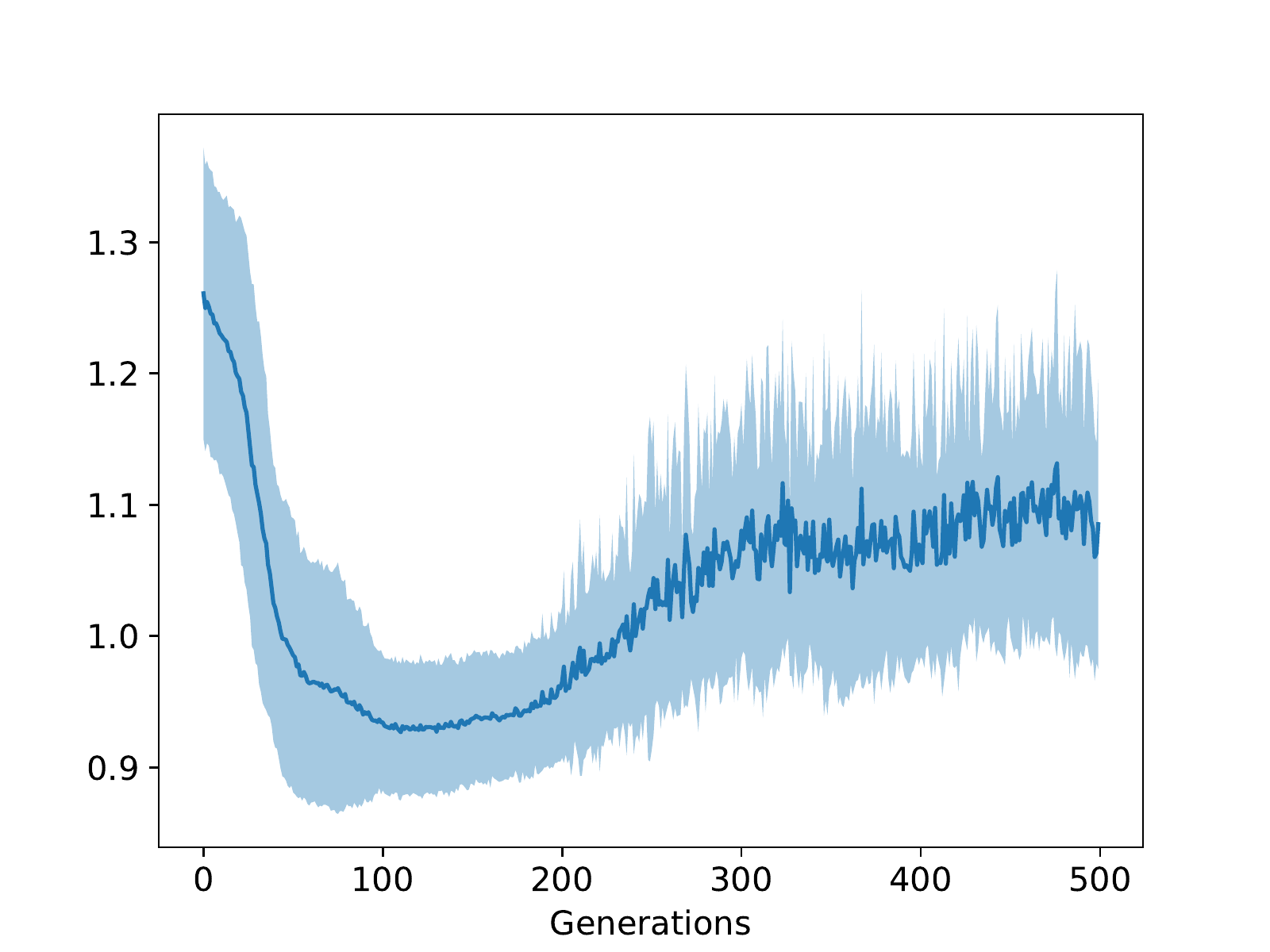}
    \caption{
      \emph{Quality predictions} on training data $\XA$.
    }
    \label{fig:RMSEFitTrain}
  \end{subfigure}
  \caption{
    Root mean squared error (with standard deviation (SD)) of different metrics on \lcsone{}'s elitist's performance, averaged over 30 random AM-Gauss problems.
  }
  \label{fig:RMSEs}
\end{figure*}

\begin{figure}[h]
  \begin{subfigure}[t]{.47\linewidth}
    \includegraphics[width=\linewidth]{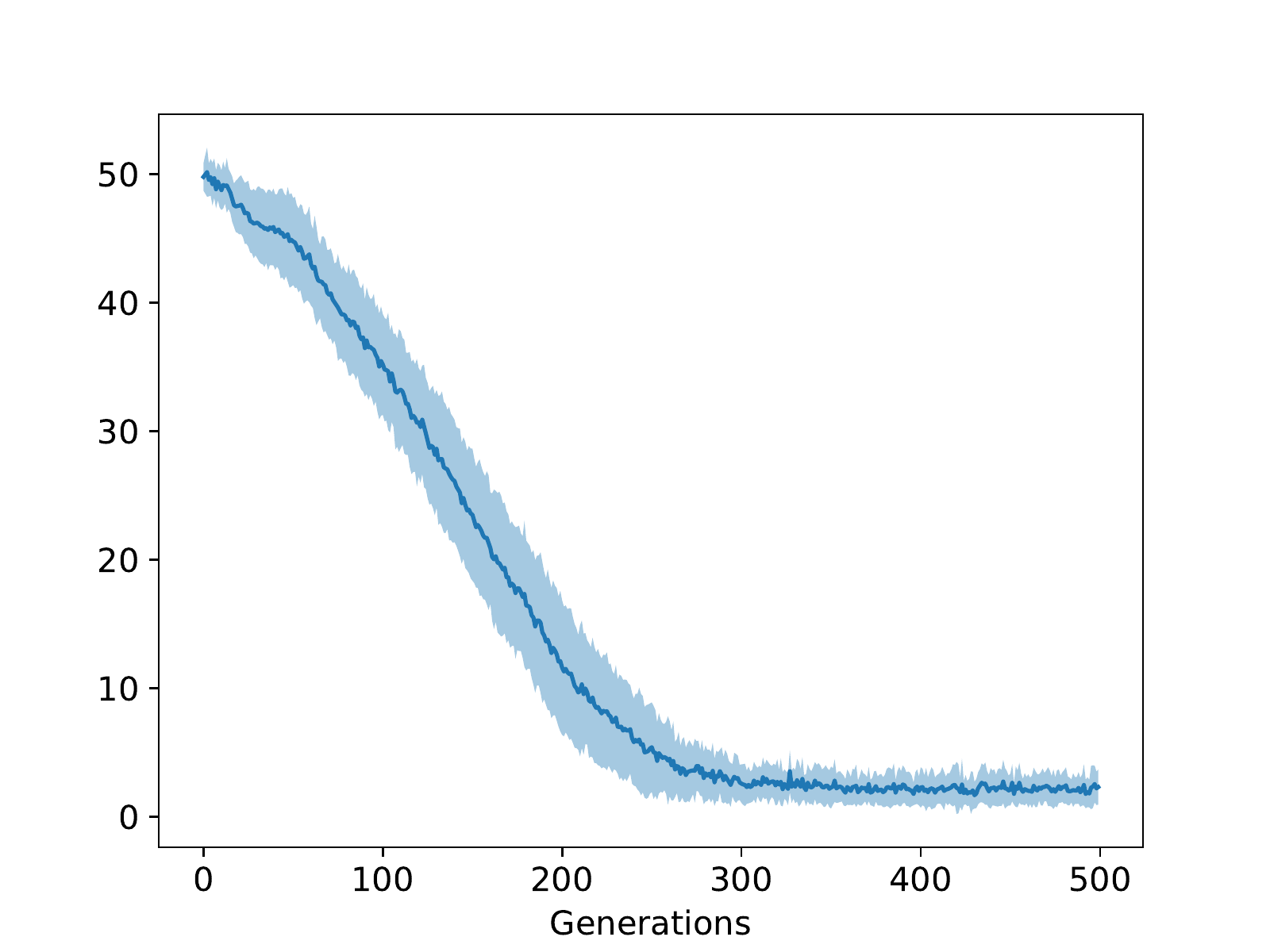}
    \caption{
      Number of classifiers in \lcsone{}'s elitist on AM-Gauss problem holdout data.
    }
    \label{fig:NClassifiers}
  \end{subfigure}
  \hfill
  \begin{subfigure}[t]{.47\linewidth}
    \includegraphics[width=\linewidth]{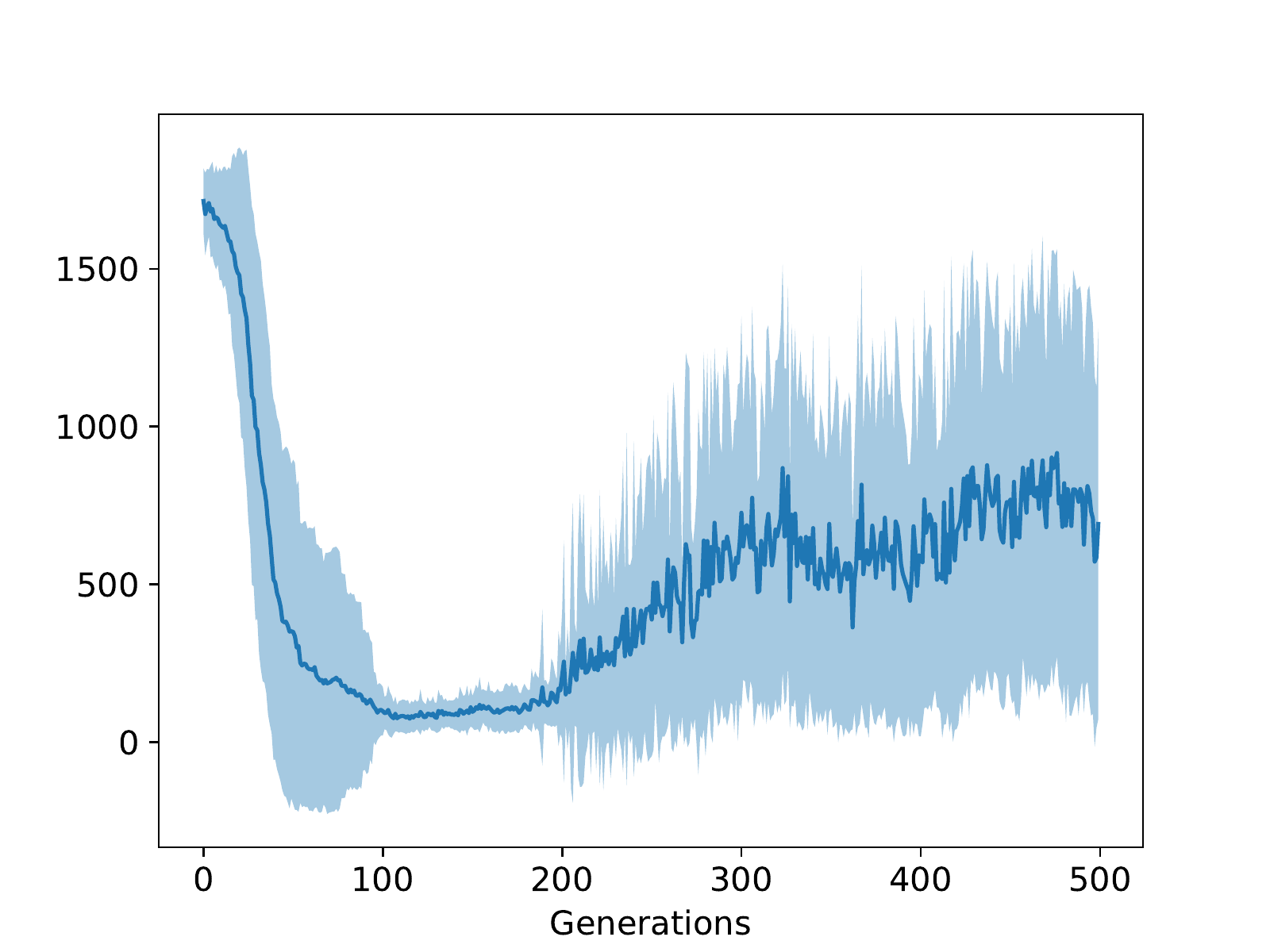}
    \caption{
      Number of unmatched examples of training data $\XA$.
    }
    \label{fig:Unmatched}
  \end{subfigure}
  \caption{Simplistic complexity and knowledge gap measurements with SD of \lcsone{}'s elitist.}
  \label{fig:Metadata}
\end{figure}

We generated 30 such functions from consecutive random seeds and used these to create 30 sets of training data for \lcsone{}.
These sets each contained 2000 examples for training (1000 training and 1000 validation examples) and 1000 examples we held out for evaluation.
On each of those data sets \lcsone{} was run once for 500 generations using all standard parameters but initial individual sizes of 50 and $k = 10^{-6}$ and then evaluated; the results are shown in Figures \ref{fig:RMSEs} and \ref{fig:Metadata}.
Note that, having 30 different functions to test \lcsone{} on leads to a better estimate of its general performance at the cost of having a higher variance of results than when repeatedly testing on a single function.

We compare \lcsone{}'s results with those achieved by a two-layer fully connected \emph{artificial neural network} (ANN) trained on identical data and functions. 
We performed simple automated architecture optimization on the ANN in terms of error during validation, determining optimal architecture for the given problems at 512 and 8 hidden cells respectively, while using ReLu activation functions twice; model complexity was not factored into the architecture optimization strategy.
We show the results of this architecture on the holdout datasets as a baseline.

It can be seen in Figure \ref{fig:RMSEFitEval} that \lcsone{}'s quality predictions' RMSE on holdout data improves rapidly over the first 50 generations and then seems to converge at around 1.02 in generation 100 which is on par with the ANN baseline.
At around generation 200, however, the error starts to increase again and later fluctuates around a value of 1.1.
The same behaviour can be observed on parametrization choices' RMSE on holdout data (Figure \ref{fig:RMSEOptimum}) although the baseline is missed on that metric.
The same can be seen, however, when looking at the quality prediction's RMSE on the training data (Figure \ref{fig:RMSEFitTrain}); this means that the problem can be detected and averted during training especially since the number of examples that are not matched by any classifier increases in a similar manner (Figure \ref{fig:Unmatched}).

When looking at the number of classifiers in \lcsone{}'s elitist (Figure \ref{fig:NClassifiers}), a steady decrease up to a convergence at only 2-4 classifiers can be observed.
By construction it is highly unlikely that the AM-Gauss problem can be solved satisfyingly by this few local models.
We tried to alleviate that problem by during mutation adding a random classifier with a probability of 0.5 (this is also used for the shown runs)---but to no avail.
It can be seen that, between generations 100 and 200, the number of classifiers lies between 13 and 35 which seems to be the sweet spot with the used hyperparameters.

The fact that after finding that sweet spot model complexity still decreases leads us to to believe that there is an issue with \lcsone{}'s fitness measure.
And indeed: It accepts individuals with slightly worse error in favour of a lower complexity (see (\ref{eq:Fitness2})), which, when applied repeatedly can result in classifier population deterioration such as the one observed.
This problem can easily be fixed by making (\ref{eq:Fitness2}) dependent on the error of the best individual ever seen.

Besides, due to the $k$ hyperparameter being more delicate than expected (see Section \ref{sec:Hyperparameters}) we assume that the value we determined for these runs was by far not optimal.

\section{Related Work}
\label{sec:relatedwork}

The \lcs{} learning system is inspired by previous research work in the field of learning classifier systems.
LCS have been applied to a diverse field of problems resulting in a diverse field of applications, e.\,g.\@ function approximation \cite{wilson2002a, urbanowicz2015b}, complex multiplexers \cite{urbanowicz2015a}, robot kinematics \cite{stalph2012, marin2011}.
LCS research can  be further divided into Michigan- and Pittsburgh-style systems, with Michigan-style systems featuring a GA operating on the level of individual rules, where the entirety of rules represents a solution to the problem, while in Pittsburgh-style (also abbreviated as Pitt-style) systems the GA operates on sets of rules, where each set represents a complete solution to the problem.

The most famous and well researched family of LCS stem from Wilson's XCS classifier system \cite{wilson1995} following the Michigan-style.
XCSR \cite{wilson2000} expanded rule representations and therefore input options from ternary to continuous by using interval predicates; a representation used by many following systems such as XCSF \cite{wilson2002a}, BioHEL \cite{bacardit2009} and of course \lcsone{}.
XCSF is an extension for XCS to perform function approximation by replacing the constant payoff prediction with a local linear model, thus performing a piecewise-linear approximation of the overall function.
The linear local models have subsequently been replaced by more complex models, such as higher order polynomials \cite{lanzi2005}, interpolation \cite{stein2018} or neural networks \cite{lanzi2006}.

Pittsburgh-style systems perform well when following a supervised learning setup and have usually been applied to classification/data mining problems.
GALE \cite{llora2001} performed data mining for various classification tasks such as the detection of breast cancer, solving multiplexers and the classification of irises.
NAX \cite{llora2007} has been applied to the diagnosis of prostate cancer without human input.
GAssist \cite{bacardit2004, franco2013} was build for supervised learning of classification tasks and uses a standard GA to evolve rules basing on GABIL \cite{dejong1991}.
As typical for Pittsburgh-style LCS, individuals consist of a set of rules that represent a complete problem solution of variable length.
The solution returned at the end of training is the highest fitness individual.
The basic system uses discrete inputs and continuous inputs get discretised into dynamically generated micro-intervals.
Not covered samples get predicted as a default class whose samples were not used for training.

A more recent example of Pittsburgh-style systems for classification on discretised data is EDARIC \cite{santu2014}. It is designed to deal with both over-fitting and class-imbalance by evolving populations for each class separately and using ensemble techniques for unknown samples. Generalisation is achieved by starting from maximally specific rules and gradually deleting constraints on less relevant input attributes. It was shown to perform well when compared to XCS, decision trees and GAssist for a number of classification datasets.

BioHEL \cite{bacardit2009, franco2013}, a descendent of GAssist leaving the traditional Pittsburgh-style behind, is using an iterative rule learning approach to learn continuous and discrete attributes for bioinformatic datasets. 
It uses XCSR's hyper-rectangle representation for continuous inputs and GABILs representation for discrete inputs. Fitness is based on GAssist's fitness function with the addition of including coverage of rules. It uses the default classification mechanism of GAssist.

In reinforcement learning real world applications can often not be represented by discrete actions.
Thus, the field of continuous actions in XCS has found much research.
While the problem described in section 1 is not understood to be in a reinforcement learning context and would only be of a single step nature, the optimal parametrization follows a similar design principle to multi-dimensional continuous actions.
Wilson proposed three architectures \cite{wilson2007}: IAL, a second XCSF interpolating the choices of the decision making XCSF, CAC, an actor-critic approach, and GCS, where a continuous action is aggregated with the input and a function of both is learned using XCSF.
In GCS the optimal action is the action maximizing the learned function.
The general approach of GCS is thus related to \lcs{} with the important distinction that GCS matches on both action and state.

Tran et al.\@ introduce XCSFCA \cite{tran2007} as another way to deal with continuous actions by computing the action directly from the input.
XCSFCA approximates a function $(\XX, (\XX \to \AA)) \to \RR$ which, due to currying, corresponds to $\XX \to ((\XX \to \AA) \to \RR)$.
Since $A$ is never a domain, XCSFCA only learns exactly one (the best regarding the quality measure) action $\amaxp (x)$ for each $x \in X$.
That optimal $\amaxp (x)$ is modelled using a linear model which is optimized separately using (1+1)-ES.
\lcs{} instead learns $(\XX, \AA) \to \RR$ which can be written as $\XX \to (\AA \to \RR)$.
Thus it is able to approximate the quality $q(x, a)$ of every possible actions $a \in \AA$, which is far more informative than only getting the best possible action, as the resulting action quality function $\qp$ could be analysed at will afterwards.
The same argument can be made for all systems using computed actions.
Besides the above, the structure of Tran et al.'s system is deliberately close to the one of XCS(F).

Howard et. al.\@ \cite{howard2009} expanded on the idea of computed actions in XCSF by using a neural network to determine both matching and actions from the given inputs.
Iqbal et al.\@ \cite{iqbal2012} also dealt with continuous actions by computing them in XCSRCFA, where the action is represented by a code fragment of a two branches deep binary tree that is evolved when creating new classifiers, similar to genetic programming.
Naqvi and Browne \cite{naqvi2016} incorporated this approach to solve symbolic regression problems.

Hashemnia et al.\@ \cite{hashemnia2017} incorporate continuous actions into XCSR to balance an unmanned bicycle in simulation. However, to choose an action to execute they discretise the actions, determine a discrete set by a fitness-weighted roulette wheel selection mechanism and choose the continuous action of the fittest classifier within the determined set.

\section{Future Work}
\label{sec:FutureWork}

Given that \lcsone{} was deliberately designed to implement \lcs{} while being as simplistic as possible there are numerous additions that can and will be made.
Some of them are already in the works, such as an expansion to hyper-ellipsoid conditions \cite{butz2005c} and testing of different polynomial and non-polynomial local models such as sine, exponential and radial basis functions (e.\,g.\@ similar to the Gaussians already used as a testbed for \lcsone{}).
An expansion to neural networks seems plausible as well \cite{lanzi2006}, although, in order to keep the degree of explainability high, they have to be kept as simplistic as possible.

Further, an investigation of a heterogeneous model landscape seems desirable, as some parts of a function might be harder to approximate even with very specific classifier conditions while others can be described linearly with ease.
\lcs{} keeps model complexity low while keeping performance high (see Section \ref{sec:ModelStructure}), we can thus expect that simpler models will be chosen where appropriate, for example, by including model type into the mutation operator.

In \lcsone{}, the model structure is evolved by a GA, however, other algorithms such as CRO \cite{lam2010} are capable of improving an underlying model structure.
Although this would arguably make \lcs{} no longer a strict representative of Pittsburgh-style LCS, as this name is strongly linked to GAs, an investigation seems appropriate.
Both improvement techniques for GAs (e.\,g.\@ n-point-crossover, different crossover and mutation rates) and different model structure optimizers should thus be investigated.
As explainability is a key feature of \lcs{}, we will compare it with other machine learning techniques commonly seen as explainable such as decision trees.

In \lcsone{} the search space is not partitioned in $\AA$ to increase explainability, as understanding a singular function is much easier than understanding a combination---possibly including mixing models---of multiple heterogeneous functions. 
Note, that understanding a single function even of low order polynomials is non-trivial.
A hierarchical approach where multiple classifiers will be located within a classifier matching a situation will be investigated in terms of performance and critically evaluated with regard to explainability.

Finally, \lcsone{} should be applied to real industrial datasets.

\section{Conclusions}

We introduced the \lcs{} learning system, a general accuracy-based Pittsburgh-style LCS architecture for supervised learning on continuous multi\--dimension\-al decision problems.
We laid the ground work for further investigation of this system by clearly defining parametrization optimization, the task \lcs{} is primarily meant to perform, describing the overall architecture and providing a first, deliberately simplistic, implementation (\lcsone{}) of it.
Said implementation was evaluated on a problem from the continuous-action LCS literature as well as on an abstract, simplified model of an industrial FDM manufacturing process.
\lcsone{} has some shortcomings but these can be attributed to its simplicity.
The overall approach shows a lot of prospect.

\begin{acks}
This work was in part supported by the German Federal Ministry for Economic Affairs and Energy (BMWi).
\end{acks}

\bibliographystyle{ACM-Reference-Format}
\bibliography{References}

\end{document}